# Lost but not erased: Finding traces of a forgotten language in neural speech models

Peter Plantinga, Charlotte Moore, Peter W. Donhauser, Krista Byers-Heinlein, & Denise Klein

## Abstract

International adoptees retain phonological traces of a birth language they can no longer speak or comprehend, a persistence typically attributed to a biologically-timed critical period. We asked whether it could instead reflect the ordinary dynamics of learning, using automatic speech recognition models that simulate the international adoptee experience without maturational confounds. Models were trained on one language and then abruptly switched to a second. We found that traces of the first language persisted throughout second-language training, but mainly in the lowest, pre-phonemic layers. These traces were functional, as models with early exposure re-learned their lost first language 14% faster than naive models; this advantage held even against models adopted early from a related language and disappeared when the earliest layers were substituted from a non-adopted model. We argue that these critical-period effects reflect entrenchment of foundational representations rather than a maturational loss of plasticity, and that experience plays a central role in critical periods in language acquisition.

## Introduction

International adoptees offer a natural experiment for understanding the effects of early language exposure on subsequent brain development. These individuals abruptly lose access to their pre-adoption language early in life, then rapidly acquire their post-adoption language, typically catching up with their peers within a few years[1]. In adulthood, they exhibit linguistic performance on par with lifelong monolingual speakers of the post-adoption language, with no conscious recollection of their pre-adoption language[2].

This apparent completeness of adoptees' birth language loss is at odds with critical periods in language acquisition[3,4], often described as a maturationally-based window of heightened neural plasticity in which early experience should leave a lasting trace. Indeed, careful experimentation has revealed that the pre-adoption language is not fully lost. For example, Mandarin-to-French international adoptees show brain activation patterns that better match speakers with continuous Mandarin access than French monolinguals[5,6]. Adoptees also show faster relearning of the pre-adoption language: Korean-born international adoptees raised in Sweden and the Netherlands have shown accelerated relearning of Korean phonological contrasts[2,7]. This faster relearning is a modern instance of the classic savings effect first described by Ebbinghaus[8]: material that appears to have been forgotten is reacquired more quickly than it was initially learned. Together, these findings show that traces of the phonology (sounds) of the pre-adoption language persist despite years of post-adoption language exposure and no conscious memory of the lost language.

Yet, other research has demonstrated that adoptee advantages do not extend to all aspects of language. Korean-Swedish adoptees' advantage on the pre-adoption language held only for phonology, and adoptees scored below non-adopted Swedish speakers on higher-complexity tasks like Korean grammaticality judgments[2]. A similar asymmetry appears in artificial systems, even though they lack biological maturation: deep neural networks initially trained on blurred images showed persistent deficits in classification accuracy, but not when the images were instead flipped vertically, a manipulation that preserves fine-grained spatial statistics[9]. These results suggest that entrenchment of early-learned information is not uniform across a learning system but is concentrated in its foundational, low-level representations.

Complex perceptual domains are represented hierarchically in both biological and artificial systems: successive processing stages extract low-level features, such as visual textures or acoustic spectro-temporal edges, and progressively combine them into higher-level categories such as objects or words[10,11]. This hierarchy could explain the asymmetry observed above. Because higher-level categories are built from lower-level features, revising a lower-level representation mid-stream could destabilize everything computed on top of them, whereas a high-level category could be revised with comparatively little cost, since few downstream computations depend on it.

This hierarchical explanation has direct empirical support in adult language acquisition. Finn and colleagues found that learning a constructed language with unfamiliar sounds required adults to recruit more general auditory processing networks rather than relying cleanly on native-language networks, concluding that adult language learning difficulties stem in part from the rigidity of phonetic neural scaffolding[12]. Under this view, low-level representations of sounds resist mid-stream changes, likely because revising these foundational structures risks destabilizing the higher-level representations that depend on them. We propose that critical-period effects can arise from this pressure to preserve a stable statistical foundation, without requiring an appeal to biological maturation.

This proposal remains untested, however, because no existing study has isolated the mechanism: human studies show the phenomenon but cannot rule out biological maturation as its cause, while artificial-network studies have focused on behavioral metrics in computer vision[9,13] or high-level text modeling[14] without localizing the critical-period effects. By contrast, this study begins to isolate the mechanisms of critical-period effects by simulating the adoptee experience in deep speech models and investigating the internal representations of the speech encoder. Using layer-wise representational analyses, we directly test where in the network language traces persist, providing mechanical evidence in favor of statistical explanations for critical-period effects, rather than biologically-timed maturational schedules.

More specifically, we have used state-of-the-art automatic speech recognition (ASR) models[15], which must learn phonological structure from raw speech input much as human infants do. Because the system was artificial, we could precisely control factors such as the learning rate throughout training, eliminating the maturational confound that limits interpretation of human studies. We previously validated this modeling framework for bilingual phonological development, demonstrating the emergence of parallel phonological systems in simultaneous bilinguals and cross-language transfer in sequential bilinguals[16].

Here we extended this approach to model international adoptees: each model was first trained on a pre-adoption language ($L_{pre}$) before training was abruptly switched to a post-adoption language ($L_{post}$), mimicking the international adoptee experience. Whereas previous studies of humans examined adoptees initially exposed to a tonal language that later acquired a non-tonal language, we trained adoptee models on non-tonal languages that are more closely related (French, German, and English), thus examining the generalizability of international adoptee effects.

Our adoptee models ($M_a$) reproduced the behavioral profile of human international adoptees, rapidly losing proficiency in $L_{pre}$, while acquiring native-like proficiency in $L_{post}$. They also showed similar advantages in identifying $L_{pre}$ phonemes that human international adoptees show in forced-choice tasks[7]. Examining their representational geometry revealed that traces of $L_{pre}$ persisted selectively in the model's earliest layers, mirroring the neural traces of a lost birth language documented in human international adoptees[5,6], and this persistence carried functional consequences for how quickly the lost language could be relearned, echoing relearning advantages documented behaviorally in human adoptees[2,7]. Varying the duration of initial training revealed an optimal window for trace preservation: with extended pre-adoption exposure, representational traces peaked before plateauing and slowly declining. Finally, swapping the earliest layers with a non-adopted model removed a significant fraction of the adoptee re-learning advantage. Together, our findings offer a mechanistic account of how early experience can leave lasting, consequential structure in a learning system without any change in its capacity to learn.

## Analytic Approach and Results

Our overall method is presented in Figure 1, consisting of four steps: 1. model training, 2. similarity matrix generation, 3. representational similarity analysis (RSA), and 4. trajectory analysis.

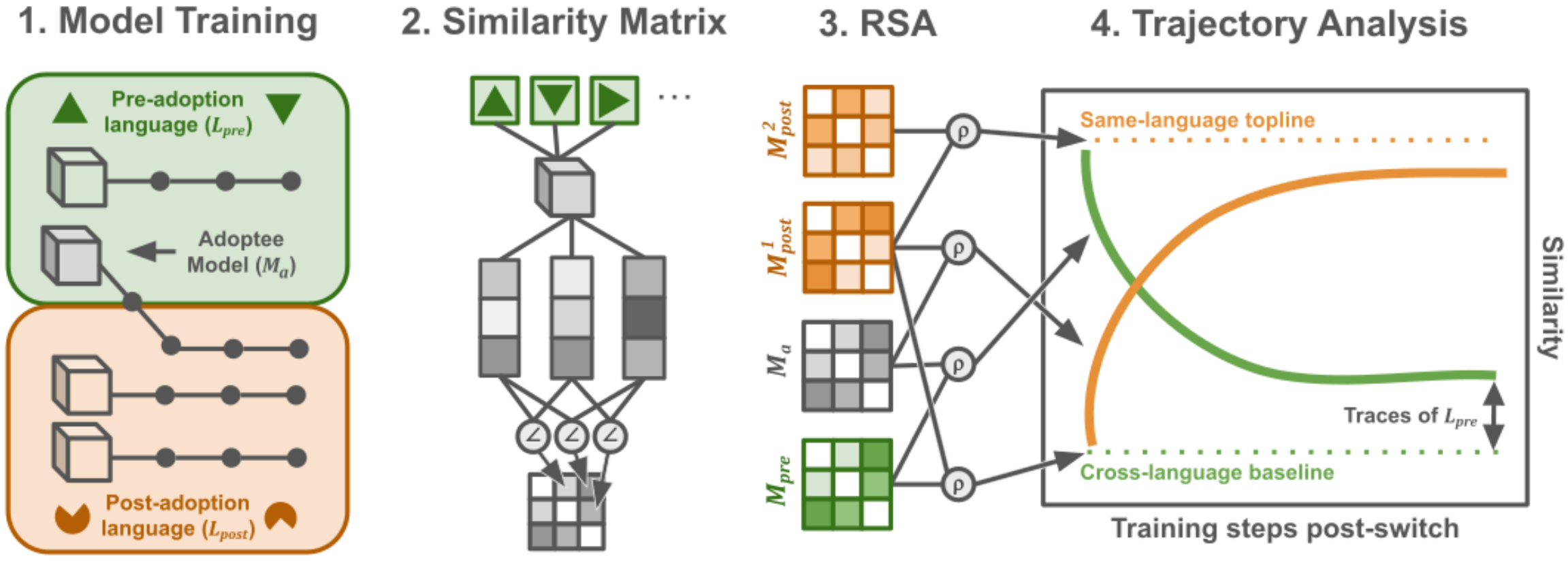

**Figure 1. Schematic of the modeling and analysis pipeline.** We trained automatic speech recognition (ASR) models under an abrupt language switch and found traces of the lost language by comparing adoptee models to single-language models using representational similarity analysis (RSA). **Step 1 (model training)**: We trained models under different language schedules: adoptee models ($M_a$) were trained on a pre-adoption language ($L_{pre}$) and then a post-adoption language ($L_{post}$), and single-language models ($M_{pre}$, $M_{post}$) were trained on one language each. Dots on the training trajectory mark saved checkpoints. **Step 2 (similarity matrix generation)**: Within a model, and at a given layer, we computed pairwise cosine similarity ($\angle$) of representations for a fixed set

of held-out $L_{pre}$ samples to form a representational similarity matrix. **Step 3 (RSA)**: We compared the similarity matrices of two models using Spearman's rank correlation (ρ), quantifying how similarly they organize the same samples. **Step 4 (trajectory analysis)**: We tracked RSA over the course of training to determine whether traces of the pre-adoption language remained stable throughout extended post-adoption training.

Across all experiments, we trained encoder–decoder ASR models on French, English, and German speech and abruptly switched their training language to simulate international adoption (see Methods). Importantly, we did not explicitly train the model on the phonology of any language, similar to other models where phonology emerged from word-level training[16,17]. We switched the language after a consistent number of updates across experiments: after models reached basic proficiency in its pre-adoption language, measured by word error rate (WER, the proportion of words transcribed incorrectly; lower scores are better) but long before final convergence. For most experiments we use 8.4k updates, after which French WER was 34.2%, German 13.8%, and English 27.6%. Throughout, we compared these adoptee models ($M_a$) against baseline models trained on a single language ($M_{pre}$ and $M_{post}$).

## Behavioral Links

We first asked whether our artificial adoptee models match the behavioral profile of human international adoptees. Throughout training, we tracked proficiency in each language using next-token prediction accuracy: the model's ability to predict the next linguistic unit of an utterance from the audio and the preceding text, where each unit is a subword token from a data-driven tokenizer. We used next-token accuracy rather than WER here because it provides a fast-to-compute and low-variance view of accuracy, rather than relying on slow search algorithms that can fail for entire utterances well into training. To track the loss of $L_{pre}$ specifically, we paired the adoptee model's evolving encoder with the original pre-adoption decoder (from $M_{pre}$) after the switch, letting us measure how accessible $L_{pre}$ remained as $L_{post}$ training proceeded.

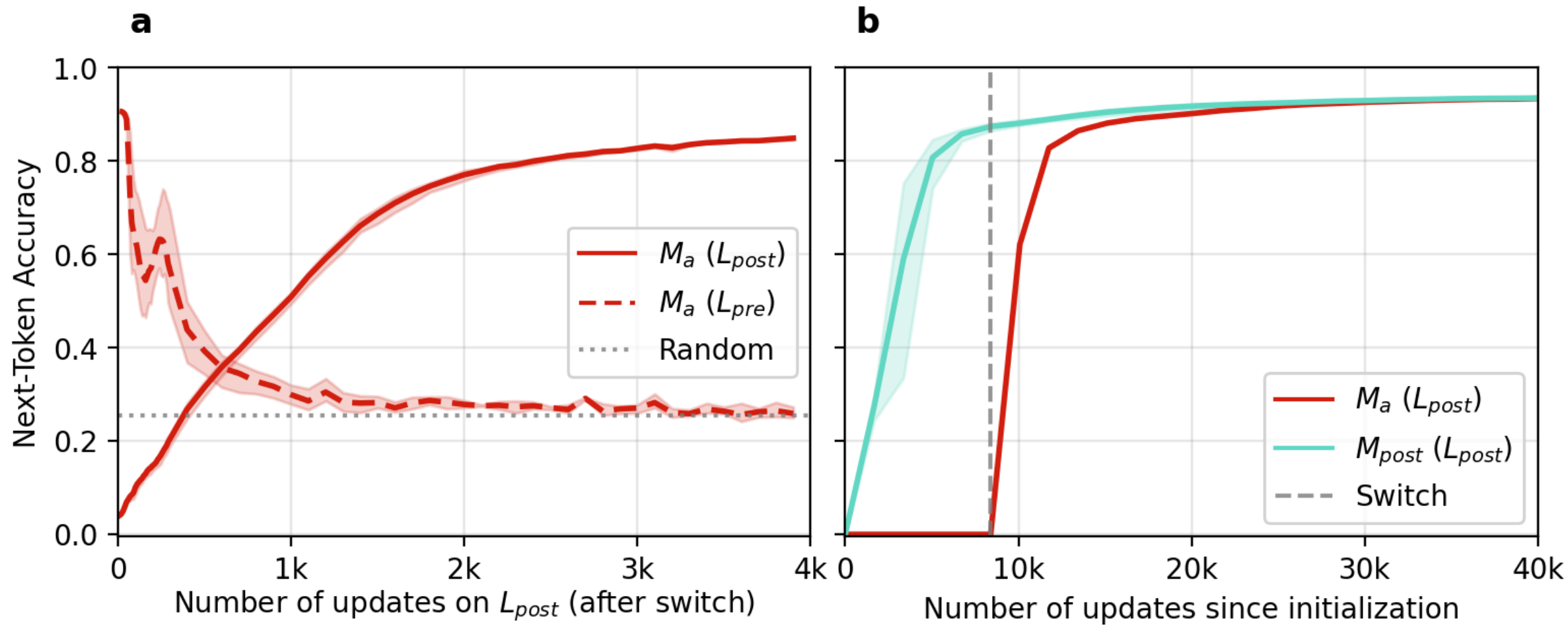

**Figure 2. The adoptee model rapidly loses its pre-adoption language while gaining a new one, mirroring human international adoptees.** Figures show next-token identification accuracy of an adoptee model ($M_a$) at two timescales. In both panels, the y-axis is next-token prediction accuracy, the model's ability to predict the next subword token from the audio and the preceding text. Lines show the mean across 3 seeds, bands around all lines indicate 95% CIs. **Panel a:** Shortly after the switch, $M_a$ accuracy on the pre-adoption language ($L_{pre}$ ; German, in blue) fell sharply, while its accuracy on the post-adoption language ($L_{post}$; French, in red) rose over the

same period. The x-axis shows the number of training updates in the post-adoption language. **Panel b:** Over the full course of training, $M_a$ (red) reached the same $L_{post}$ (French) accuracy as compared to a model only trained on French ($M_{post}$, yellow). The x-axis shows the number of updates since training began in any language; the vertical dotted line marks the onset of $L_{post}$ training for $M_a$.

The adoptee model showed a sharp decline in its $L_{pre}$ (German) recognition accuracy, but gained proficiency in $L_{post}$ (French) over the same timespan, mirroring the typical pattern in human international adoptees who rapidly acquire the adopted language. Recognition accuracy in $L_{pre}$ declined over 4000 updates to 25.9% (see Figure 2), approaching the accuracy obtained when the decoder was given randomly shuffled inputs instead of a real audio representation (25.4%), indicating that the decoder acted as a next-token predictor (i.e. ignored the audio entirely). At the same time, $L_{post}$ accuracy rose rapidly, reaching 80% after only 3000 updates post-switch, 30% fewer steps than it took $M_{post}$ to reach the same level. With the total number of updates (on any language) held fixed, $M_a$ and $M_{post}$ converged to less than 0.1% accuracy difference (93.6% vs. 93.7%). This pattern aligns with observed behavior in human international adoptees, who lose conscious access to their birth language and in most cases[18] gain proficiency similar to their monolingual peers in the adopted language[1,19].

A second well-documented aspect of human international adoptee behavior is an enhanced ability to discriminate phonemic contrasts from their birth language[2]. We asked whether $M_a$ shows an analogous advantage in its representations of $L_{pre}$ phonemes. Because human experiments ask participants to identify phonemes with little explicit feedback[20], we used a linear probe[21], trained for 300 steps. We froze each model and trained a simple linear classifier on the internal hidden states of each layer to predict the corresponding time-aligned phoneme (see Methods). Above-baseline probe accuracy therefore reflects phonemic information already present in the frozen representations rather than new phonemic learning.

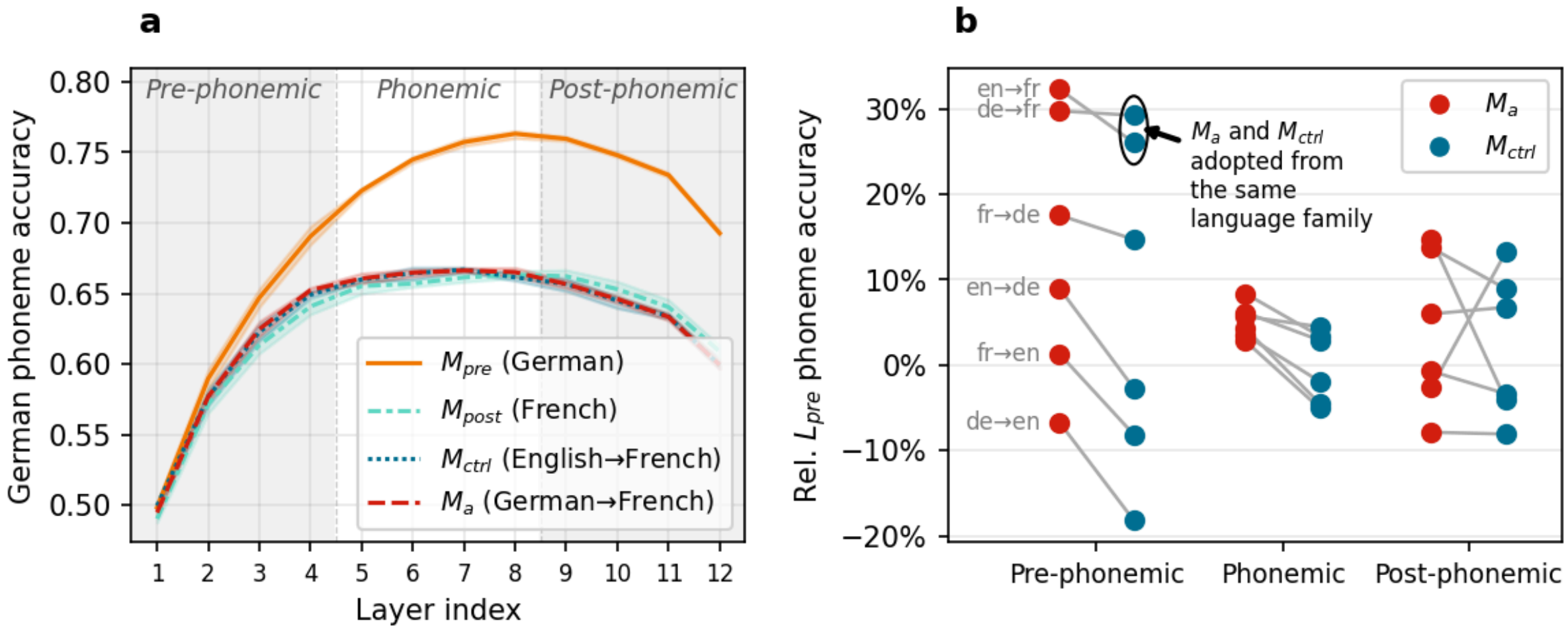

**Figure 3. The adoptee model retained a phoneme-discrimination advantage for its pre-adoption language in early processing (pre-phonemic) layers.** Linear phoneme probe classification accuracy over the layers of the encoder. **Panel a:** Absolute accuracy of a linear probe trained to classify pre-adoption phonemes ($L_{pre}$; German) across representations from single-language models (German, French) and adoptee models ($M_a$ = German→French; $M_{ctrl}$ = English→French). The x-axis is the encoder layer index (1-12); probe accuracy is the percentage of frames correctly labeled with their aligned phoneme. Shaded regions denote functional layer groupings (*Pre-phonemic*, *Phonemic*, and *Post-phonemic*). Confidence bands represent 95% intervals across model training runs. **Panel b:** Normalized probe accuracy for $M_a$ and $M_{ctrl}$ aggregated by layer group across six

language switch conditions. Accuracy is expressed relative to single-language benchmarks where $M_{pre}$ = 100% (maximum) and $M_{post}$ = 0% (baseline). Connected points represent a family of models like the example given in panel a, evaluated on phonemes from the pre-adoption language of the adoptee model.

Figure 3a shows the probe's accuracy across encoder layers for an example language switch direction (i.e. German → French). For all four compared model families, probe accuracy peaked in the middle layers (5-8), establishing them as the models' "phonemic layers." Both $M_a$ and $M_{ctrl}$ showed an advantage over the $M_{post}$ baseline at the earlier pre-phonemic layers ($p$ = 0.0011 and $p$ = 0.019 respectively, one-sided Mann-Whitney $U$), as well as at the phonemic layers for $M_a$ but not $M_{ctrl}$ ($p$ = 0.013 and $p$ = 0.057), and not for the post-phonemic layers either ($p$ > 0.9 for both). This pattern extended across language pairs (Figure 3b) where it can be seen that the advantage of $M_a$ over $M_{post}$ was largest at pre-phonemic layers and shrank as depth increased. Furthermore, the probe was more accurate for $M_a$ than $M_{ctrl}$ at the pre-phonemic and phonemic layers ($p$ = 0.016 for both, one-sided Wilcoxon signed-rank) but not at post-phonemic layers ($p$ = 0.28).

These data reveal two distinct sources of early-exposure advantage. First, the strong phonetic identification ability of $M_{ctrl}$ in Figure 3a and the highlighted portion of Figure 3b is most naturally read as positive forward transfer, in which early exposure to any Germanic phonology deposits acoustic-phonetic features that also serve languages in the same family, such as spectral edges or formant ranges. Second, the consistent $M_a$-over-$M_{ctrl}$ increment from Figure 3b indicates an advantage attributable to $L_{pre}$ exposure specifically, over and above the family-level transfer that $M_{ctrl}$ also enjoys. Together these results suggest that the adoptee models retain low-level, language-family-specific patterns from $L_{pre}$ most strongly at the foundational levels of representation, rather than intact higher-level phonemic categories.

## Representational Traces of Lost Language

Neuroimaging studies have shown identifiable traces of early language experience when stimuli from a lost language are presented to international adoptees[5]. To look for an analogous trace in our models, we used representational similarity analysis (RSA)[22] to compare how models with different language histories organize their internal representations. RSA is well suited here since it captures whether two models represent the same stimuli similarly without requiring that the representations be in the same space, handling geometric differences that arise due to random initialization and data ordering (e.g. rotations and translations). We computed RSA using 500 held-out $L_{pre}$ speech samples, comparing each model's representational geometry layer by layer (see Methods).

Figure 4a shows RSA scores across model pairs for models with different training schedules. We bracketed the range with two bounds: an upper bound from pairs of same-language models trained on different seeds (high similarity) and a lower, cross-language bound from $M_{post}$ - $M_{pre}$ pairs (low similarity). These baselines represent the level of architectural similarity we would expect from two monolinguals who either speak the same language or two different languages. Against these bounds, we tracked the similarity of $M_a$ to both $M_{pre}$ and $M_{post}$ over the course of $L_{post}$ training.

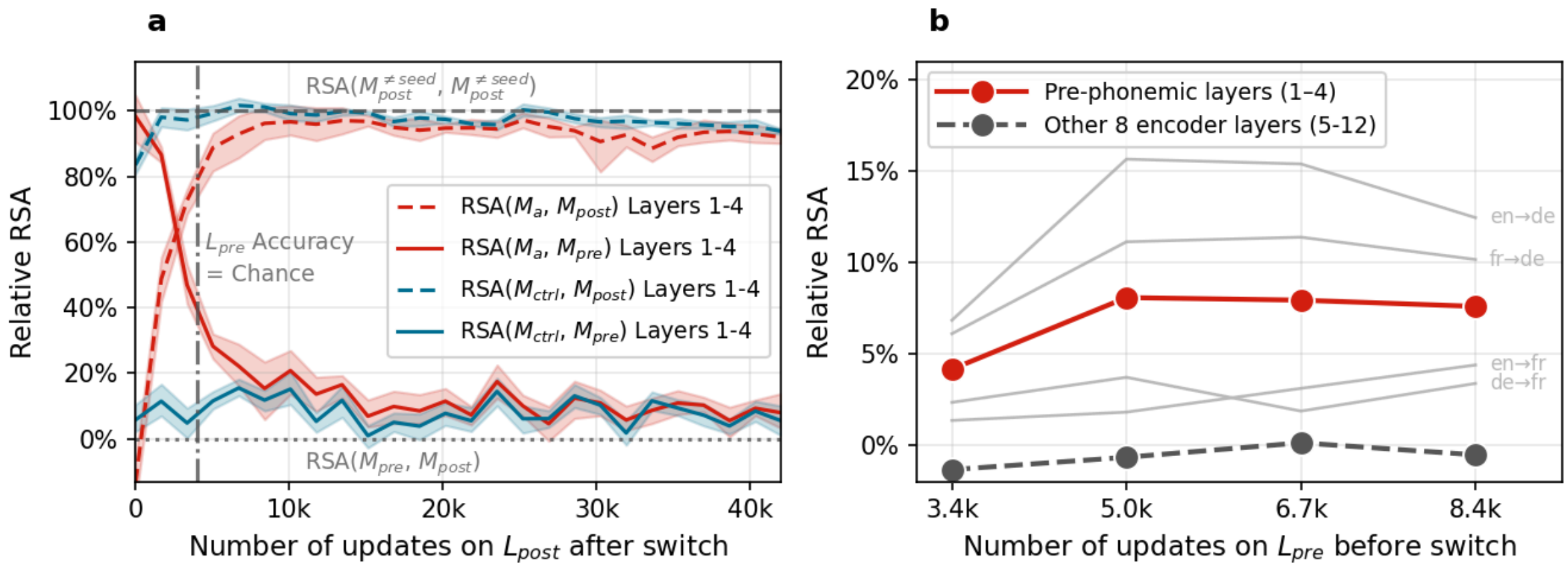

**Figure 4. Representational traces of the lost language that persist after behavioral loss are concentrated in the lowest layers, and peak during an optimal pre-training window.** Representational similarity analysis (RSA) of adoptee models ($M_a$) against single-language models, computed on held-out pre-adoption language ($L_{pre}$) samples. Confidence bands show 95% confidence intervals across 3 seeds. **Panel a:** For a German → French adoptee model given 8.4k steps of $L_{pre}$ training, the similarity of $M_a$ and $M_{ctrl}$ and single-language $M_{pre}$ and $M_{post}$ models over the course of post-adoption training ($L_{post}$), averaged across the pre-phonemic layers (1-4). The control model was an English → French adoptee model. The x-axis is updates of $L_{post}$ training. Horizontal grey lines mark similarity scores for two monolingual models trained on the same language (high similarity, dashed line) and a lower, cross-language similarity baseline (low-similarity, dotted line). $M_a$'s similarity to $M_{post}$ approaches this upper bound, whereas its similarity to $M_{pre}$ stabilized above the cross-language baseline, indicating a persistent trace of the lost language. **Panel b:** The representational trace, defined as $M_a$'s similarity to $M_{pre}$ above the cross-language baseline, as a function of the number of $L_{pre}$ training steps. The solid red line shows the trace for the pre-phonemic layers (1-4), the average of the four language switch conditions shown in light grey; the heavy dashed grey line shows the average for these four language switch conditions across the other encoder layers (5-12), showing that the trace is concentrated in the lowest layers.

Immediately after language switch, $M_a$ representations rapidly approached $M_{post}$ representations, confirming robust adaptation to the new language. The crucial finding is that well after accuracy on $L_{pre}$ had fallen to chance (4k steps, see Figure 2), $M_a$ looked more similar to $M_{pre}$ than a $M_{post}$ model without $L_{pre}$ exposure. Moreover, this residual similarity did not fade, as even after new-language training equivalent to thousands of hours of experience, the representational geometry stabilized rather than continuing to converge on $M_{post}$. As can be seen in Figure 4a, after around 15k updates the RSA($M_a$, $M_{pre}$) curve flattened, averaging 8.3% (95% CI = 2.9% to 13.7%) over the last 5k steps. The control model's score was also significantly more similar than chance (mean 6.3%, 95% CI 2.4% to 10.2%). As with the phoneme probe, this significant control trace indicates that part of the representational similarity reflects language family-level transfer, i.e. Germanic phonetic information rather than German transfer specifically; $M_a$'s numerically higher score is consistent with an additional $L_{pre}$-specific increment, but the overlapping intervals mean RSA does not, on its own, cleanly separate the two components. That separation is achieved by the relearning experiment below.

To determine where and when these traces formed, we evaluated representational similarity across encoder layers and initial exposure lengths (Figure 4b). As in the phoneme probe, traces were sharply localized to the earliest, pre-phonemic layers (layers 1–4), reaching a relative RSA of 4% to 8%, whereas the remaining 8 higher-level encoder layers showed virtually no residual trace (remaining near zero baseline). Crucially, the magnitude of these pre-phonemic traces did not scale continuously with $L_{pre}$ training. Instead, relative RSA nearly doubled between 3.4k and 5.0k pretraining updates (rising from ~4% to ~8%), reached a clear peak around 5.0k–6.7k steps, and began a slow tail-off by 8.4k steps. This non-linear pattern was consistent across 3 out of 4 individual language switch pairs (Figure 4b, light grey lines), while the English → French $M_a$ models show consistently stronger representational traces with more pre-training.

Taken together, these results show that representational traces of the lost language persist through extended new-language training, peaking after a limited pre-training duration rather than accumulating indefinitely.

## Savings on Relearning

Finally, we asked whether these traces serve a functional purpose or are merely residual imprints from early language experiences. We investigated this by testing how quickly adoptee models could relearn their lost language.

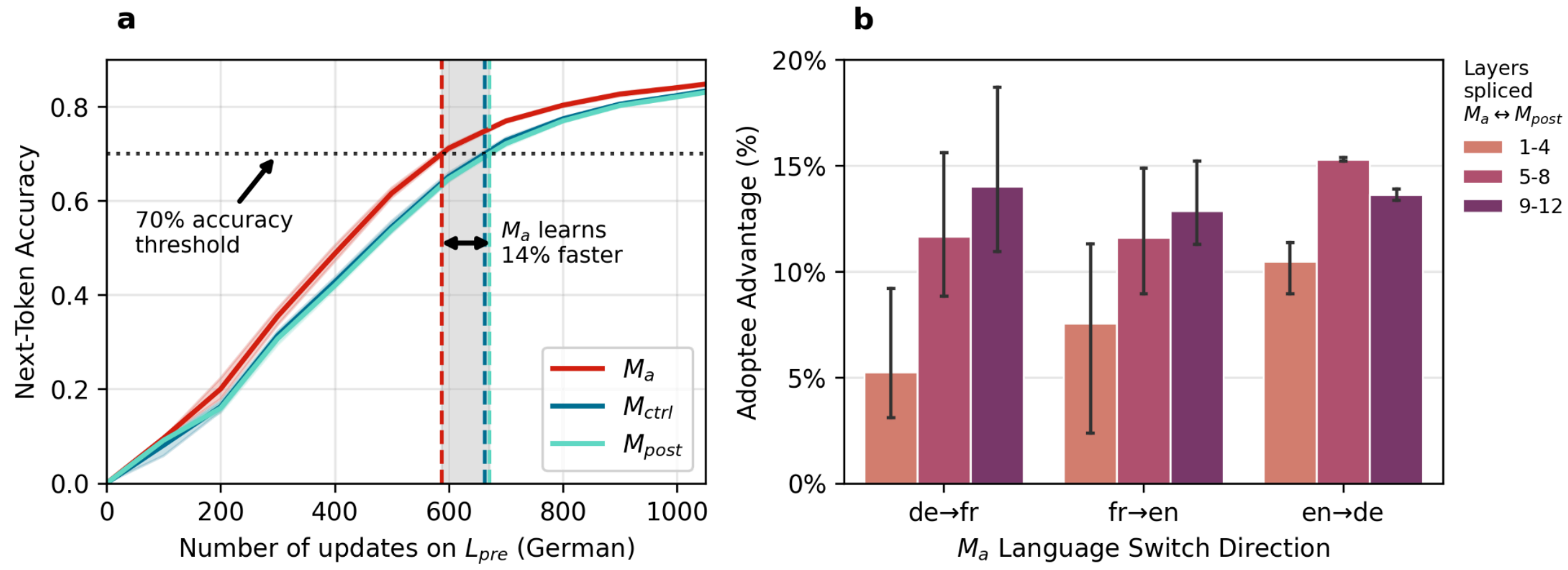

**Figure 5. The adoptee model relearned its lost language faster than control models, but lost this advantage when the earliest layers are artificially excised.** Relearning of pre-adoption language (here, German) by models with various language backgrounds. **Panel a:** Next token accuracy scores for next-token prediction as a function of number of learning updates on $L_{pre}$ (here, German). X-axis shows the number of $L_{pre}$ updates the model has experienced in the new learning schedule. Y-axis shows next-token accuracy. Red represents $M_a$, models exposed to German for 8.4k updates and then abruptly switched to French. Blue represents $M_{ctrl}$, adoptee models whose previous language experience do not include German (instead, English). Cyan represents $M_{post}$ models that have also never been exposed to German. Horizontal dotted line marks the fixed accuracy threshold (70%) used to define a target for comparison; vertical lines show the point at which each model crosses the threshold. Color bands represent 95% CIs. **Panel b:** Relative advantage of $M_a$ over $M_{post}$ in learning speed when portions of the encoder have been exchanged between the two models. Speed is measured in steps to 70% accuracy when relearning the pre-adoption language $L_{pre}$.

As shown in Figure 5a, $M_a$ relearned $L_{pre}$ faster than both $M_{post}$ models and a control model ($M_{ctrl}$) that was adopted from a third language, reaching a fixed 70% accuracy threshold in 14.3% fewer steps than $M_{post}$ (95% CI 12.3% to 16.2%) and 12.8% fewer steps than $M_{ctrl}$ (95% CI 10.9% to 14.8%). Because $M_{ctrl}$ shares $M_a$'s adoption history but lacks early $L_{pre}$ experience specifically, this advantage cannot be attributed to a generic benefit of having learned another language early on; it depends on prior exposure to $L_{pre}$. This is the savings effect[8], previously demonstrated in deep networks for computer vision[23], and indicates that relearning is facilitated by the retained phonological traces of the lost language.

To test what portion of the network drives the relearning advantage, we conducted a mechanistic localization experiment with pre-phonemic, phonemic, and post-phonemic portions of the encoder, as shown in Figure 5b. We spliced different sets of layers between $M_a$ and $M_{post}$, and tested the effect on relearning speed, isolating the effect of the representations from any general effect of the splicing operation by only comparing the relative performance between the models. Splicing the pre-phonemic layers consistently resulted in the smallest gap between the models, confirming that these layers serve a critical role in supporting the relearning of the pre-adoption language. The differential contribution of early versus late network layers suggests that the adoptee advantage is driven primarily by representations analogous to early-developing auditory–phonological systems rather than later-emerging higher-order linguistic representations.

# General Discussion

Human language learners show a surprising combination of flexibility and entrenchment in language acquisition, as exemplified by populations like international adoptees who demonstrate lasting neural traces of a forgotten language without any overt knowledge of it[5,6]. We used a neural network model trained on raw human speech data to investigate the neural traces left behind by previous language experience. Our results reproduced the same effects observed in international adoptees: models showed a complete behavioral loss of the pre-adoption language and eventual native-like proficiency in the post-adoption language. Of note, representational traces of the pre-adoption language in the model's geometry persisted, particularly at the lowest levels of representation. These traces in turn conferred a functional advantage, accelerating re-learning of the pre-adoption language phonemic contrasts relative to models with equivalent language experience but no early exposure.

These results support a hierarchical account in which early learning persists only when it establishes foundational representations. Consistent with this view, the persistent representational trace and its causal contribution to relearning was confined to the lowest, pre-phonemic layers, while the decodable phoneme advantage extended only slightly higher, into the phonemic layers, and vanished entirely at the post-phonemic and higher levels. We propose that early-formed low-level representations become entrenched and are reused, in modified form, when input changes, whereas higher-level representations built upon them are more readily overwritten. This interpretation is consistent with evidence that prior learning alters the neural mechanisms supporting subsequent learning: for example, rats trained on one spatial navigation task no longer require the same brain plasticity to learn a second task[24], and early language experience shapes neural responses to later phonological working memory tasks[6]. Rather than simply demonstrating that early language experience leaves a lasting trace, our findings identify the type of representation that persists and the stage of processing at which it exerts its influence. Although deep neural network layers are not direct analogues of brain regions, they provide a computational means of localizing the relearning advantage to early processing stages, suggesting that early language exposure leaves durable low-level representations that facilitate later relearning.

A central theme across our analyses is that behavioral performance underestimates what a learning system retains. Prior work exploring critical period effects for artificial learners has largely tracked behavioral trajectories following early deprivation [9]. By instead tracking internal speech representations after a sudden language change, we observed structures that model behavior did not reveal. The representations of our adoptee model always remained distinguishable from those of a monolingual of either language, despite near-identical behavioral proficiency (93.6% versus 93.7% accuracy). This finding bolsters the relationship between our model and human adoptees, who show neural traces even when phonemic discrimination behaviour shows no advantage[5,6]. This distinction may also explain why phonemic discrimination tasks used with human Korean-Swedish adoptees [2] detect a birth-language advantage at all: such tasks can be completed using any acoustic information that distinguishes the relevant sounds, including pre-phonemic cues. These tasks may therefore draw on the same low-level traces we find in our models, rather than on intact phonemic-level representations per se. Moreover, we found advantages for English-French adoptee models later tested on German, suggesting that in humans, low-level phonetic encoding skills should transfer across closely related languages[25], which would be consistent with the broader second language phonology literature[26,27]. Techniques that are sensitive to lower-level auditory encoding (e.g. frequency-following responses) may be well suited to test these predictions.

While our model demonstrates several connections between speech models and human international adoptee learners, several limitations should be noted. First, our models do not develop like humans do. The network's size, depth and learning rate were determined a priori rather than arrived at through maturation. We therefore cannot speak to how biological maturation might interact with these learning dynamics in humans. Our results constrain what a maturational account must explain, rather than eliminating a role for biology. Second, we directly tested low-level phonological processing, while our account predicts that higher-level representations such as grammar should show weaker latent preservation. An important future direction will be to integrate these aspects of language into a single artificial language model.

There are other, more practical limitations. Our models underwent a total and abrupt cessation of $L_{pre}$ input after the language switch. In most real-world cases of language attrition, including but not limited to international adoption, speakers continue to have some contact with their birth language, whether through occasional conversations, media, daily routines, or internal experiences such as dreams or recollections[28]. Exploring the effects of small, punctuated episodes of language re-exposure on model behavior and representations could strengthen the connection to human experience and help determine the minimum amount of continued input needed to maintain or strengthen latent representations. Finally, we did not exhaustively explore the space of modelling decisions, including network size and data scale and set of languages. We note, however, that the central phenomenon does not appear to be architecture- or task-specific as parallel results have been observed in convolutional vision models[9]. Systematically mapping how these effects vary with scale, objective, architecture, and languages remains an important direction for consolidating this evidence base.

Our results sharpen the question of the role of biology in development, rather than diminishing it. Because experience is intrinsic to every learning process and can by itself generate critical period effects, the experiential account should serve as the null hypothesis. On this view, a maturational claim about any given effect should have to show that the effect exceeds what learning dynamics alone produce[29]. Our findings give that null hypothesis concrete empirical force in a naturalistic domain, showing that the lasting effects of early language exposure observed in international adoptees need not be attributed entirely to a biologically-determined critical period, because a system with fixed plasticity already reproduces them. Any biological account of these effects cannot simply point to their persistence as evidence of a maturational window: it must specify what maturation adds beyond learning dynamics alone.

More broadly, our findings reframe what early language exposure accomplishes. Even when it leaves no measurable skill, early exposure can create durable latent structures that persist long after exposure has ceased and provide a basis for future learning. Because this structure is a product of learning dynamics rather than biological maturation, it may be possible to shape it through learning conditions, including the order in which material is presented. This idea is already exploited in machine learning as curriculum learning[30], and is relevant to educational programs such as early immersion and heritage-language maintenance. More fundamentally, our results suggest that the entrenchment of early learning, long taken as a signature of biological maturation, may instead reflect a general property of how learning systems build representations from the bottom up, with foundational structure becoming fixed as later learning comes to depend on it. Critical period effects, in this view, can arise from the order in which experience is organized over time, without requiring a biological window.

# Methods

## Model Training

All models were trained and tested on spectrograms derived from the Common Voice dataset version 22.0[31], a publicly available repository of human speech from thousands of speakers across dozens of languages. We extracted about 1,000 hours from three of the CommonVoice languages with the most data: English, French, and German, and held out 2% (20 hours) for both validation and test sets. For each language in our dataset, we trained a separate SentencePiece unigram tokenizer[32] with a vocabulary size of 5,120 tokens, chosen to balance dictionary size and encoding length.

On this data, we trained ASR models using a state-of-the-art off-the-shelf CommonVoice ASR recipe from the SpeechBrain toolkit[33], which follows an encoder-decoder formula with Conformer architecture, consisting of 12 encoder and 6 decoder layers for a total of 109.3 million parameters. The encoder ingests 80-dimensional mel-spectral features via an intervening CNN to reduce dimensionality along both time and frequency by a factor of 4, and outputs a multidimensional representation of the utterance for further processing. The decoder module receives the encoded representation, along with its own previous predictions, to transform the audio into words and sentences in an autoregressive manner. The recipe achieved near state-of-the-art performance for the CommonVoice dataset when no pretraining was used, getting a WER of 6.17% on our German test set, and 9.97% on our French test set.

One crucial detail that we changed from the off-the-shelf recipe is removing late-training learning rate decay. While standard recipes for ASR include a warmup period and a cooldown period, we included only the linear warmup period (12k updates), needed for convergence, but excluded any kind of cooldown period in middle or later stages of training. We did this to ensure that the findings of neural traces cannot be attributed to a kind of "freezing" effect, where rapid learning rate decay removes the potential of later updates to affect the geometry enough to remove traces of the early learning experiences. This does come at a small performance cost, as adding one epoch of training with learning rate cooldown to the end of training improved WER by close to 1% absolute (6.17% → 5.34% for German), but this ensured our results were generalizable effects of the learning process itself, and not a byproduct of a specific learning rate schedule. This mirrors the results of Achille and colleagues[9] who found that the sensitive period effects were reduced but not eliminated when learning rate cooldown was removed.

To simulate the adoptee experience, we trained each model on a pre-adoption language ($L_{pre}$) and then abruptly switched the training data and tokenizer to a post-adoption language ($L_{post}$) after 2-5 epochs (3.4k-8.4k updates). When switching the language, we kept all trained parameters of both the encoder and decoder for the new language training, changing only the tokenizer and data. This approach retains the greatest possible chance of saving representational geometry across the model's representational hierarchy.

For each language pair and pretraining length we trained at least three models, each with a different random weight initialization and data ordering. Confidence bands in the figures represent 95% CIs over seeds, computed using a non-parametric bootstrapping algorithm. Statistical tests are over random seeds except for one test over language switch directions in the phonemic linear probe. False discovery rate correction (Benjamini-Hochberg) applied to the p-values listed in the paper does not change any conclusion at a 0.05 significance level.

## Representational Similarity Analysis

Once the training was complete, the second step was to construct a representational similarity matrix for each model over a set of 500 stimuli randomly selected from the held-out data from $L_{pre}$. Choosing the pre-adoption language allowed us to see if samples from the forgotten language were still able to activate sub-networks of the adoptee model above the level of a model that had never heard utterances from $L_{pre}$ before.

For each of the 500 speech stimuli from $L_{pre},$ we constructed a single-vector representation of the sample by averaging over time the internal model representations for a given layer when presented with the sample. Then, we compared all pairs of stimulus vector representations using cosine similarity to construct the similarity matrix. The diagonal and replicated half of the matrix were discarded before correlating matrices to avoid artificially inflating similarity scores.

In the third step, we compared similarity matrices using Spearman's rank test to estimate the degree to which two models hold similar or different internal representations for the given set of stimuli. The full set of comparisons we conducted are as follows:

1. **RSA($M_{post}, M_{post}$)** — The maximum expected correlation comes from separate models that are trained on the same language: we compared pairs of models with different random initializations—but both trained using $L_{post}$—to provide a topline comparison for whether the adoptee model fully adapts to the second domain.

2. **RSA($M_{pre}$, $M_{post}$)** —The minimum expected correlation comes from two models that have entirely non-overlapping training data coming from separate domains. These models can still exhibit some similarity from overlap in the task being performed: for example, some phonemes sound very similar across languages.
3. **RSA($M_a$, $M_{pre}$)** —The question we were investigating was whether the adoptee models' similarity with models from the pre-adoption domain would remain above a model that had never seen data from that first domain. We compared the correlation here against the RSA($M_{pre}$, $M_{post}$) correlation score listed above to determine if higher-than expected correlation could be found, a result we term "representational traces" of the first seen language.
4. **RSA($M_a$, $M_{post}$)** — Due to the recent experience of both $M_a$ and $M_{post}$ on the same language, RSA($M_a$, $M_{post}$) should nearly reach the topline RSA($M_{post}$, $M_{post}$) level, but due to the distinct early language experiences of $M_a$ some representational differences could remain, which we expected to be of roughly the same magnitude as the RSA($M_a$, $M_{pre}$) score from the baseline.

Finally, we calculated the amount of representational traces remaining and a 95% CI around this value by averaging the RSA scores over the last 5k steps to arrive at a stable estimate of representational traces per random initial seed (5 for $M_a$ and 7 for $M_{post}$), using a student's t-distribution for the small-sample CI estimate.

For the experiment on the duration of pretraining, we use a slightly different formulation of representational traces based on a headroom-normalized comparison against only the baseline without consideration for the topline placement (i.e. (RSA($M_a$, $M_{pre}$) - RSA($M_{pre}$, $M_{post}$) / (1 - RSA($M_{pre}$, $M_{post}$)). This is due to some language switches having a very small baseline-topline gap in the early layers causing the variance to be too high to extract recognizable patterns.

## Phonemic Linear Probe

To determine if the adoptee model retained a phoneme-learning advantage in a similar manner as international adoptees, we trained a linear probe (i.e. an affine transformation) to predict, for each frame, the corresponding aligned phoneme. To conduct this experiment, we relied on the Montreal Forced Aligner (MFA)[34] toolkit with MFA-trained language-specific acoustic models (version 3.0.0) to generate estimates of phoneme timings based on a comparison of the raw audio with the ground-truth transcripts. We limited the phoneme inventory to a set of 51 overlapping phonemes by collapsing allophones over the three languages. For each frame, we then selected the phoneme that occupied the majority of the frame as the label during linear probe training.

For each model, we conducted the probe experiment by first freezing all the weights of the model encoder, and applying a separate linear probe to the outputs of each layer, trained with a cross-entropy loss. The probe trained quickly with only 300 updates at a constant 0.001 learning rate needed to reach an accuracy plateau. A total of 5 models out of 93 were removed as outliers due to very poor phonemic probe performance (<50% accuracy at the middle layer), likely due to an inability to distinguish one or more crucial common phonemes.

The statistical tests determining if linear probe scores exceeded the baseline were conducted over 6 seeds for the adoptee model and 4 seeds for the control model (2 excluded as outliers), by averaging the normalized scores at each layer into an overall score for each model and layer bin combination. Then a non-parametric one-sided Mann-Whitney *U* test was performed to test if the model scores exceeded the baseline. We further aggregated scores over language switch directions, resulting in six conditions per model (adoptee or control) per layer bin. Then we tested whether the direction of the effect over the 6 language pairs was significant using a one-sided Wilcoxon signed-rank test.

### Savings on Relearning

To test for a "savings effect" in all of our models, we (re-)trained our fully-converged models on $L_{pre}$. We used an accelerated warmup schedule (1000 updates to full learning rate) as the models were already well-structured and did not require a long warmup to converge. We measured validation performance every 100 steps and used linear interpolation between the two nearest points to find the number of training steps needed to reach our arbitrary accuracy threshold.

To localize the relearning speed advantage in the adoptee model, we spliced a sequence of layers from each $M_a$ with a corresponding $M_{post}$ and vice-versa. After splicing, we froze the entire model except the first and last fully-connected portions of the spliced layers, and adapt on 2000 steps of the $L_{post}$ to compensate for rotations or translations of the representations between models. Then, language re-learning proceeded as before. Confidence intervals of the adoptee advantage were computed using bootstrapping over model seeds.

## Data Availability

The CommonVoice dataset is available through the Mozilla Foundation.

## Code Availability

Code is available at https://github.com/pplantinga/bilingual_networks.

## Author Information

These authors contributed equally: Peter Plantinga, Charlotte Moore.

These authors jointly supervised this work: Krista Byers-Heinlein, Denise Klein.

### Affiliations

**McGill University Department of Neurology and Neurosurgery**
Peter Plantinga, Charlotte Moore, Denise Klein

**Concordia University Department of Psychology**
Charlotte Moore, Krista Byers-Heinlein

**Ernst Strüngmann Institut**
Peter Donhauser

**Centre for Research on the Brain, Language, and Music**
Peter Plantinga, Charlotte Moore, Krista Byers-Heinlein, Denise Klein

**Mila Quebec Artificial Intelligence Institute**
Peter Plantinga

### Contributions

All authors were involved in conceptual development. P.P. and C.M. designed the experimental protocols. P.P. and P.D. wrote the experimental code. P.P. ran the experiments and generated the figures. K.B.-H. and D.K. provided expert guidance and feedback. All authors participated in writing and editing the manuscript.

### Corresponding Author

Peter Plantinga